\documentclass[letterpaper, 10 pt, conference]{ieeeconf}  

\IEEEoverridecommandlockouts                              

\title{\LARGE \bf
Acquisition of interpretable domain information \\during brain MR image harmonization for content-based image retrieval
}

\author{Keima Abe$^{1}$ , Hayato Muraki$^{1}$ , Shuhei Tomoshige$^{1}$ , Kenichi Oishi$^{2}$ , Hitoshi Iyatomi$^{1}$
\\for the Alzheimer’s Disease Neuroimaging Initiative*
\thanks{*Data used in preparation of this article were obtained from the Alzheimer’s Disease Neuroimaging Initiative (ADNI) database (adni.loni.usc.edu). As such, the investigators within the ADNI contributed to the design and implementation of ADNI and/or provided data but did not participate in analysis or writing of this report. A complete listing of ADNI investigators can be found at: \protect\url{http://adni.loni.usc.edu/wp-content/uploads/how_to_apply/ADNI_Acknowledgement_List.pdf}
}
\thanks{This work was not supported by any organization}
\thanks{$^{1}$ Dept. of Applied Informatics, Graduate School of Science and Engineering, Hosei University, 3-7-2 Kajino Koganei, Tokyo, 184-8584,
Japan
}
\thanks{$^{2}$ Dept. of Radiology and Radiological Science, Johns Hopkins Medicine, 225 Traylor Building,
720 Rutland Ave.Baltimore, MD 21205, USA}
}

\usepackage{graphicx}	
\usepackage{amsmath}	
\usepackage{amssymb}	
\usepackage{booktabs}
\usepackage{times}
\usepackage{microtype}
\usepackage{epsfig}
\usepackage{caption}
\usepackage{float}
\usepackage{placeins}
\usepackage{color, colortbl}
\usepackage{stfloats}
\let\labelindent\relax
\usepackage{enumitem}
\usepackage{tabularx}
\usepackage{xstring}
\usepackage{multirow}
\usepackage{xspace}
\usepackage{url}
\usepackage{subcaption}
\usepackage{color}
\usepackage{xcolor}
\usepackage[hang,flushmargin]{footmisc}
\usepackage{pifont}
\usepackage{listings}
\usepackage{cite}
\usepackage{makecell}

\newif\ifreview 
\newif\ifarxiv 
\newif\ifcamera 
\newif\ifrebuttal 

\ifcamera \usepackage[accsupp]{axessibility} \fi
\ifarxiv  \fi

\newcommand{\cmark}{\textcolor{black}{\checkmark}} 
\usepackage{caption}  
\usepackage{bm}  
\usepackage{setspace}
\usepackage[font=small,skip=1pt]{caption} 
\begin{document}

\maketitle
\thispagestyle{empty}
\pagestyle{empty}

\begin{abstract}
Medical images like MR scans often show domain shifts across imaging sites due to scanner and protocol differences, which degrade machine learning performance in tasks such as disease classification. Domain harmonization is thus a critical research focus. Recent approaches encode brain images $\boldsymbol{x}$ into a low-dimensional latent space $\boldsymbol{z}$, then disentangle it into $\boldsymbol{z_u}$ (domain-invariant) and $\boldsymbol{z_d}$ (domain-specific), achieving strong results. However, these methods often lack interpretability—an essential requirement in medical applications—leaving practical issues unresolved. We propose Pseudo-Linear-Style Encoder Adversarial Domain Adaptation (PL-SE-ADA), a general framework for domain harmonization and interpretable representation learning that preserves disease-relevant information in brain MR images. PL-SE-ADA includes two encoders $f_E$ and $f_{SE}$ to extract $\boldsymbol{z_u}$ and $\boldsymbol{z_d}$, a decoder to reconstruct the image $f_D$, and a domain predictor $g_D$. Beyond adversarial training between the encoder and domain predictor, the model learns to reconstruct the input image $\boldsymbol{x}$ by summing reconstructions from $\boldsymbol{z_u}$ and $\boldsymbol{z_d}$, ensuring both harmonization and informativeness. Compared to prior methods, PL-SE-ADA achieves equal or better performance in image reconstruction, disease classification, and domain recognition. It also enables visualization of both domain-independent brain features and domain-specific components, offering high interpretability across the entire framework.
\end{abstract}
\section{Introduction}

\begin{figure*}[t]
  \centering
  \includegraphics[width=0.8\textwidth]{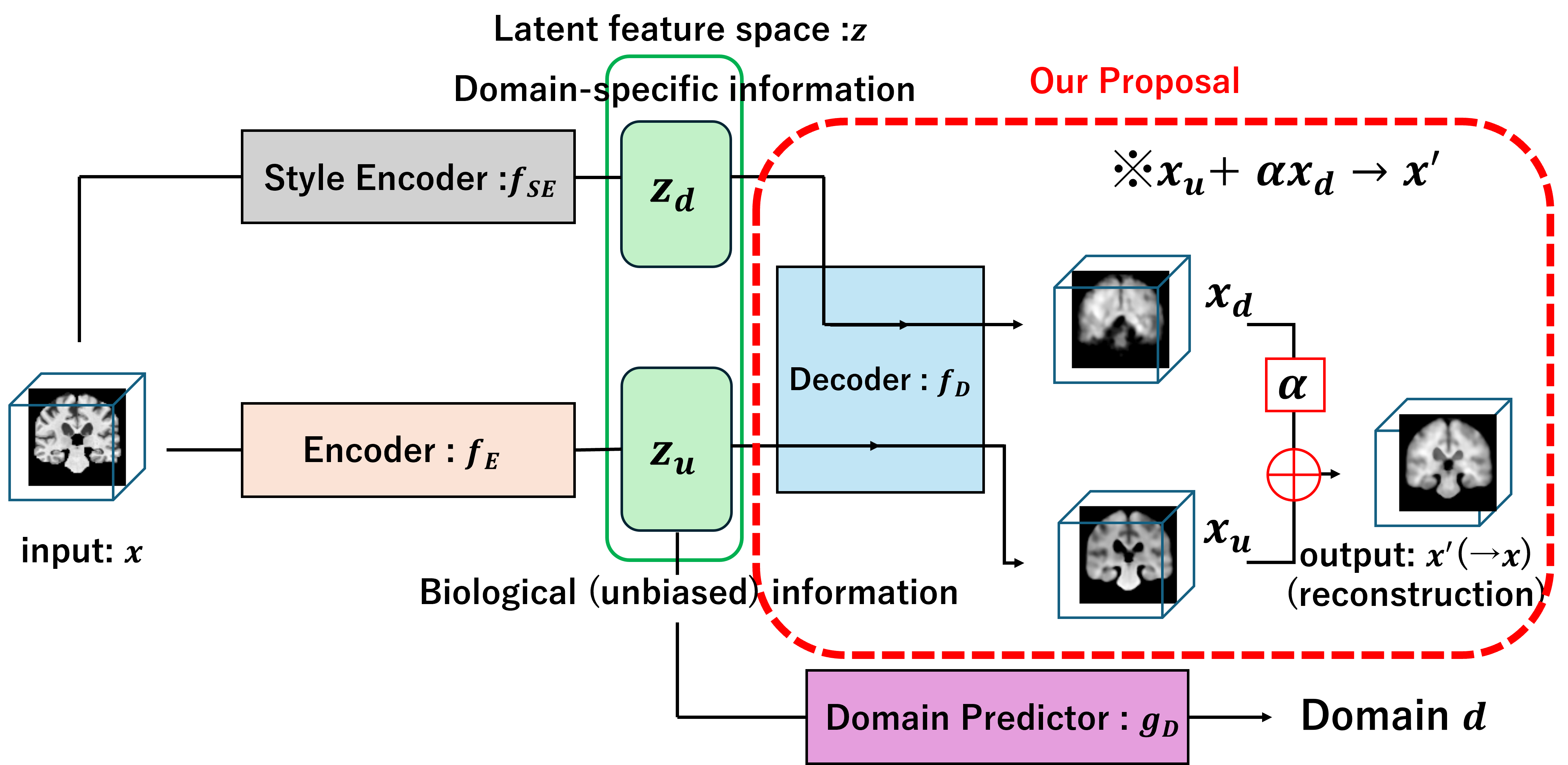}
  \caption{Architecture of PL-SE-ADA.}
  \label{fig:model_architecture}
\end{figure*}

In neuroradiology, MR images are widely used for diagnosing neurological disorders,
and the acquired images are stored in medical image management systems~\cite{choplin1992picture} together with patient information. 
However, these images are often underutilized outside of the imaging sessions themselves, and there is growing interest in their reuse. Although keyword-based retrieval methods for finding similar past cases have been proposed, they require domain expertise and experience to specify appropriate keywords. Therefore, there is a strong demand for implementing a Content-Based Image Retrieval (CBIR) system~\cite{kumar2013content, valente2013dicoogle,arai2018significant} that enables users to search for similar cases by directly inputting an image.

To realize CBIR, it is essential to extract low-dimensional representations of each case that reflect disease-related and biological features, followed by processes such as clustering. Various methods have been proposed to achieve this goal~\cite{tommasino2023histopathological, silva2020interpretability}. Meanwhile, advances in machine learning and the increasing number of studies based on large-scale datasets collected across multiple medical centers have brought greater attention to the domain shift problem, where differences in data acquisition protocols and imaging devices across centers (i.e., domains) become evident~\cite{tzeng2015simultaneous}. This domain shift induces non-biological experimental variations that significantly affect the images and hinder machine learning tasks~\cite{onga2019efficient, clark2006impact, han2006reliability, yu2018statistical, oishi2019developmental, wachinger2021detect}. Given this context, the realization of a general-purpose CBIR system for brain MR images requires a retrieval method that remains effective across datasets acquired using different scanners and imaging protocols.

For example, Wachinger et al.~\cite{wachinger2021detect} analyzed a total of 35,320 MR images from 17 publicly available datasets and conducted a task called ``Name That Dataset'' in which the objective was to identify the dataset solely based on the image.

They reported that using only volumetric and cortical thickness features extracted from 70\% of the training data, they classified the dataset with 71.5\% accuracy. This result indicates that features such as volume and cortical thickness strongly reflect dataset-specific properties. Eliminating such biases is essential for multi-center collaborative studies, long-term follow-up research, and the construction of robust CBIR systems.

Based on previous studies, the realization of a practical CBIR system for brain MR images requires the acquisition of a low-dimensional representation that satisfies the following four criteria:

\begin{enumerate}[label=(\roman*)]
\item Fidelity – faithfully preserve salient anatomical structure and disease‑related features of the original image (minimal information loss).
\item Domain robustness – remain invariant to acquisition‑site differences such as scanner model and imaging protocol.
\item Direct usability – support similarity search directly in the latent space without auxiliary classifiers.
\item Interpretability – offer a semantically meaningful structure that clinicians can inspect and trust.
\end{enumerate}

Criterion (i) pertains to the integrity and preservation of the data. Criterion (ii) is a crucial requirement for ensuring robustness against external disturbances that are unrelated to the subject’s condition. Criterion (iii) refers to usability—specifically, the ability to perform retrieval based on the distance between low-dimensional representations without requiring an external classifier. Lastly, criterion (iv) concerns the readability and interpretability of the representation.

Arai et al. proposed a domain harmonization method in the image space called Disease-Oriented Image Embedding with Pseudo-Scanner Standardization (DI-PSS) to eliminate the effects of non-biological image variations~\cite{arai2021disease}. This method employs a 2D image transformation model based on CycleGAN to convert brain MRIs acquired from arbitrary scanners into images that simulate those taken by a reference scanner. By absorbing inter-scanner differences as a preprocessing step, and subsequently applying metric learning focused on disease-related features, this approach achieved promising results in content-based image retrieval using low-dimensional representations, fulfilling requirements (i), (ii), and (iv).

As a method addressing requirement (ii) — the removal of domain differences — techniques based on ComBat~\cite{johnson2007adjusting}, originally developed within an empirical Bayes framework to correct for technical and conditional variations in data acquisition, have been proposed and are now widely used~\cite{beer2020longitudinal}. However, ComBat is not suitable for CBIR, as it lacks essential capabilities such as compressing brain images from unseen domains and retrieving similar cases.

Dinsdale et al. addressed requirement (ii), which involves removing domain differences, by proposing a method that extracts domain-invariant low-dimensional representations using an encoder for dimensionality reduction combined with adversarial training against a domain predictor. They reported promising results with this approach~\cite{dinsdale2021deep}.
However, due to the conflicting objectives of domain harmonization and preserving comprehensive brain information in the low-dimensional representation, requirement (i) remains challenging to achieve.

Tobari et al. proposed Style Encoder Adversarial Domain Adaptation (SE-ADA), which extends the method by Dinsdale et al. toward the realization of CBIR~\cite{10.1117/12.3046652}. SE-ADA separates the original brain MR image into two components: information essential for CBIR and domain-dependent information. It achieves a balance between harmonization learning and reconstruction learning. By training on multiple domains within a single model, SE-ADA enables domain harmonization in a single training session. The low-dimensional representations acquired by this method satisfy requirements (i) and (ii), while significantly improving training efficiency. However, the low-dimensional representations obtained by SE-ADA, which include domain information, still present challenges in terms of interpretability—requirement (iv)—which is particularly critical in the medical domain.

Muraki et al. proposed Isometric Feature Embedding for CBIR (IE-CBIR), a low-dimensional representation learning method based on RaDOGAGA, which ensures isometry between real and latent spaces, and Baseline++, which enables classification based on distances in the low-dimensional space. Their method successfully satisfies requirements (i) and (iii)~\cite{muraki2024isometric}.

Although various studies have been conducted to achieve CBIR for practical medical imaging, as described above, few have successfully addressed the interpretability requirement outlined in (iv) to date.

To address this issue, we propose Pseudo-Linear-SE-ADA (PL-SE-ADA), a novel and versatile framework for CBIR that significantly enhances the interpretability of the resulting low-dimensional representations, building upon SE-ADA. While inheriting SE-ADA's advantage of separating domain information within the low-dimensional space, PL-SE-ADA further enables the visualization of both biologically essential (subject-related) information and domain-specific information as separate images—offering a level of interpretability not previously achievable. In summary, the proposed PL-SE-ADA satisfies three requirements for CBIR-related data representation: (i), (ii), and (iv).

\section{Pseudo-Linear-SE-ADA (PL-SE-ADA)}
In this paper, we propose Pseudo-Linear-SE-ADA (PL-SE-ADA), an extension of SE-ADA~\cite{10.1117/12.3046652} that enhances model interpretability while preserving effective domain harmonization and disease-relevant feature representation in brain MR images.

\subsection{Overview of PL-SE-ADA}
Fig.~\ref{fig:model_architecture} shows the architecture of PL-SE-ADA.
PL-SE-ADA shares the same architecture as SE-ADA, consisting of an image encoder ($f_E$), a style encoder ($f_{SE}$) that extracts domain-invariant ($\boldsymbol{z_u}$) and domain-specific ($\boldsymbol{z_d}$) low-dimensional features from an input image $\boldsymbol{x}$, a decoder ($f_D$), and a domain predictor ($g_D$). Although the network architecture and training strategy of PL-SE-ADA are essentially the same as those of SE-ADA, it introduces a simple yet effective image reconstruction strategy that significantly enhances model interpretability.

\subsection{Training and Behavior of PL-SE-ADA}
Both the encoders and the decoder of PL-SE-ADA are trained to ensure that the input $\boldsymbol{x}$ and its reconstruction $\boldsymbol{x'}$ are identical. Additionally, adversarial training between the encoder ($f_E$) and the domain predictor ($g_D$) disentangles the original input $\boldsymbol{x}$ into two latent representations: $\boldsymbol{z_u}$, which preserves essential biological features, and $\boldsymbol{z_d}$, which captures domain-specific information only.

The training procedure consists of the following three stages:

\begin{enumerate}
    \item Train the two encoders and the decoder ($f_E$, $f_{SE}$, $f_D$) in an end-to-end manner to minimize the reconstruction error.
    
    \item Fix $f_E$, $f_{SE}$, and $f_D$, and train the domain classifier $g_D$ to minimize the domain prediction error.
    
    \item After all models are sufficiently trained, fix $g_D$ and update $f_E$ so that $\boldsymbol{z_u}$ becomes indistinguishable across domains. This is achieved by providing uniformly distributed domain labels as targets for domain prediction, effectively enforcing the encoders to generate domain-invariant representations. In other words, even with a well-trained domain classifier, the low-dimensional representation $\boldsymbol{z_u}$ is constructed such that domain classification becomes infeasible.
\end{enumerate}
The training of the encoder $f_E$ alone and Stage 3 are alternated iteratively to reinforce both accurate reconstruction of the input image and the acquisition of domain invariance in the latent representation $\boldsymbol{z_u}$.

In SE-ADA, the reconstructed image is generated from the combined latent representation:

\begin{equation}
\bm{x}' = f_D(\bm{z_u} + \bm{z_d}).
\end{equation}
In contrast, the proposed PL-SE-ADA decodes $\boldsymbol{z_u}$ and $\boldsymbol{z_d}$ separately and linearly combines their outputs:

\begin{equation}
\bm{x}' = f_D(\bm{z_u}) + \alpha f_D(\bm{z_d}) = \bm{x_u} + \alpha \bm{x_d}.
\end{equation}

This modest yet powerful modification yields a clearer disentanglement of latent factors and markedly improves visual interpretability, revealing the respective contributions of biological and domain‑specific features. In practice, PL‑SE‑ADA can visualize both components as distinct MR images and even supports arithmetic operations between them.
\section{Experiments}

\subsection{Data Sets and Preprocessing}
In this experiment, we utilized the 3D MRI datasets ADNI2 and ADNI3, publicly available from the Alzheimer's Disease Neuroimaging Initiative (ADNI)~\footnote{\url{https://adni.loni.usc.edu/}} , with the aim of studying mild cognitive impairment and Alzheimer’s disease.
The dataset used in this study includes the following three diagnostic labels.
AD refers to subjects diagnosed with Alzheimer's disease, CN indicates healthy control subjects with normal cognitive function and MCI represents individuals with mild cognitive impairment.
For the detailed information, see www.adni-info.org.

In this study, the differences between datasets were treated as domain differences. The data used in this study is summarized in Table~\ref{tab:dataset}. To prevent subject overlap between the training and test datasets, the data was partitioned based on subject IDs. Additionally, to ensure unbiased performance evaluation, only one image per subject was used during both training and testing.

\begin{table}[t]
\centering
\caption{Breakdown of MRI image data used: numbers in parentheses are number of patients.}
\label{tab:dataset}
\resizebox{\linewidth}{!}{
\begin{tabular}{lrrrrr}
\hline
      &       & \multicolumn{1}{c}{CN}        & \multicolumn{1}{c}{AD}        & \multicolumn{1}{c}{MCI}       & \multicolumn{1}{c}{total}     \\
\hline
Train & ADNI2 & 1,903 (262) & 927 (199) & 2,702 (429) & 5,532 (743) \\
      & ADNI3 & 788 (369)  & 175 (83)   & 523 (239)  & 1,486 (628) \\
Test  & ADNI2 & 75 (75)    & 39 (39)    & 73 (73)    & 187 (187)  \\
      & ADNI3 & 91 (91)    & 22 (22)    & 45 (45)    & 158 (158) \\
\hline
\end{tabular}}
\end{table}

All brain MRI images were subjected to a unified preprocessing pipeline. First, skull stripping was performed and the brain region was extracted using OpenMAP-T1 ~\cite{nishimaki2024openmap}. Subsequently, affine registration was applied to correct positional and orientation differences among the brain images. The spatial resolution was standardized to 1 mm, and the image size was set to 160 × 224 × 160 voxels. For efficient processing with deep learning models, the images were then resampled to a spatial resolution of 2 mm, resulting in a final image size of 80 × 112 × 80 voxels.

\subsection{Model architecture}
The network architecture of each model in PL-SE-ADA is arbitrary; however, in this experiment, it was configured to be the same as SE-ADA~\cite{10.1117/12.3046652}. Specifically, $f_E$, $f_{SE}$, and $f_D$ are all constructed with 14 layers. Additionally, $g_D$ is a fully connected 3-layer MLP network with an input layer of 175 dimensions, a hidden layer with 32 nodes, and an output layer of 2 dimensions. The encoder $f_E$ outputs a 175-dimensional latent representation, denoted as $\boldsymbol{z_u}$. The supervised encoder $f_{SE}$ produces a 2-dimensional vector $\boldsymbol{z_d}'$, which is subsequently expanded to a 175-dimensional representation $\boldsymbol{z_d}$ via an affine transformation layer. It is important to note that $\boldsymbol{z_d}'$ is utilized as a representation for supervised learning of domain-specific information. For detailed information about the model, please refer to the SE-ADA paper.

\subsection{Evaluation}
To assess the effectiveness of the PL-SE-ADA framework, both qualitative and quantitative evaluations were conducted. In the qualitative evaluation, reconstructed images, images generated solely from style (domain-specific) information, and images generated solely from domain-invariant information obtained through PL-SE-ADA were visualized to confirm the influence of each component on image characteristics.

For the quantitative evaluation, we assessed the method in terms of three criteria: reconstruction error, disease classification performance, and domain classification performance.

\begin{enumerate}[label=(\roman*)] 
\item \textbf{Reconstruction error}: For each model, the similarity between the input image and the reconstructed image was measured using Root Mean Squared Error (RMSE) and Structural Similarity Index Measure (SSIM)~\cite{wang2004image}, to assess the extent of information loss.

\item \textbf{Disease classification performance}:  
Low-dimensional feature representation of brain MR images obtained with each model were used for binary classification of AD and CN subjects using a classic three-layer multi-layer perceptron (MLP) with an input layer of 175 dimensions, a hidden layer with 128 dimensions, and an output layer of 2 dimensions. The classification performance was evaluated using the macro F1-score to assess the preservation of disease-related information.
For training the MLP, AD and CN cases in the training dataset were used. Evaluation was performed on the test set. To prevent the model from overfitting to specific individuals, only one image per subject was used during training.

\item \textbf{Domain classification performance}:  
Domain classification between ADNI2 and ADNI3 was performed using the same three-layer MLP applied to  the low-dimensional representations, with macro F1-score used as the evaluation metric. A macro F1-score close to 0.5 indicates that domain-invariant features have been successfully learned, as the domain classifier is unable to distinguish between domains.
To visually assess domain harmonization, Uniform Manifold Approximation and Projection (UMAP)~\cite{mcinnes2018umap} was used to project both $\boldsymbol{z_d}$ and $\boldsymbol{z_u}$ into two-dimensional space and observe domain-wise distributions.  
To minimize variations unrelated to domain differences, the MLP was trained exclusively on CN data from the training set and evaluated on CN data from the test set. To prevent overfitting to individual subjects, only one image per subject was used during training.
\end{enumerate}

We compared the performance of PL-SE-ADA against the following six methods:

\begin{itemize} 
\item[(A)] Convolutional Autoencoder (baseline). 
\item[(B)] The baseline method with Gaussian noise (mean = 0, standard deviation = 0.1) added to the obtained feature representations (Noise). 
\item[(C)] The baseline method with domain shift correction applied to the feature representations using ComBat~\cite{johnson2007adjusting}(Combat). \item[(D)] A model that performs adversarial domain adaptation only (ADA)~\cite{dinsdale2021deep}. 
\item[(E)] Improved model by adding style encoder to ADA (SE-ADA)~\cite{10.1117/12.3046652}.
\item[(F)] The proposed method (PL-SE-ADA). Here, we used $\alpha=0.2$ based on the preliminary experiments.
\end{itemize}

\subsection{Impact of hyperparameter $\alpha$}
We compared how each result changed during the training of PL-SE-ADA by varying the value of $\alpha$ in $z_u + \alpha z_d$. The value of $\alpha$ was varied across the following settings: 0.05, 0.1, 0.2, 0.5, 1.0, and 1.5.

\section{Results and Discussion}
\subsection{Qualitative Results}
\begin{figure}[t]
  \centering
  \includegraphics[width=0.9\linewidth]{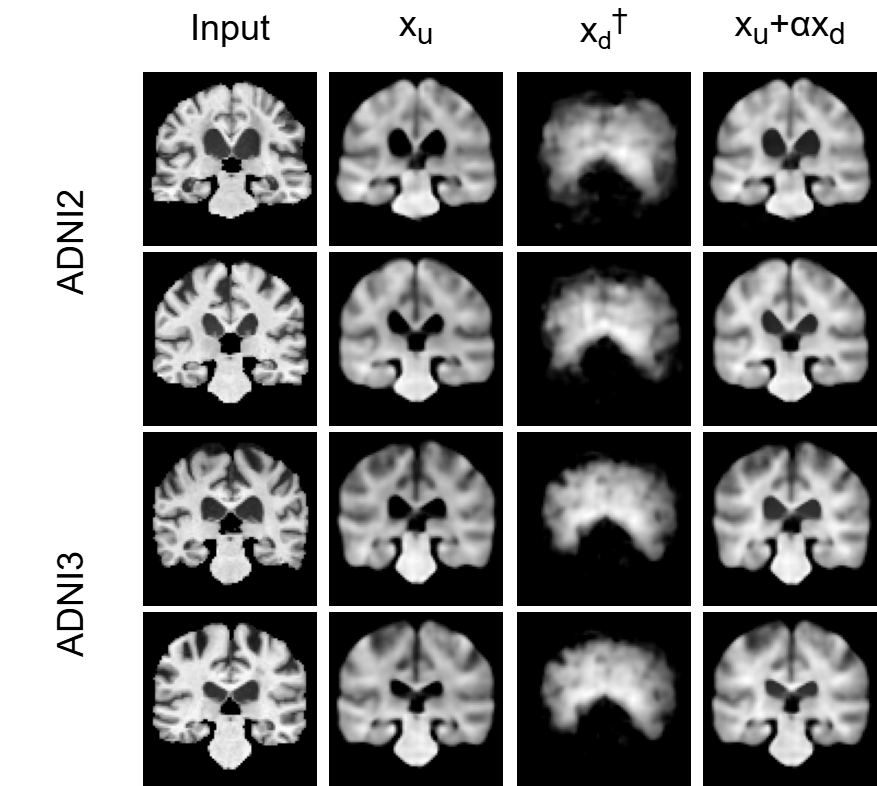}
  \caption{Examples of reconstruction with PL-SE-ADA.}
  \label{fig:reconstructions}
  \vspace{1mm}
  {\footnotesize%
  † To improve the perceptual visibility of the domain-dependent component $\boldsymbol{x_d}$, image intensity values were emphasized.%
  }
\end{figure}

Fig.~\ref{fig:reconstructions} demonstrates that both $\boldsymbol{z_u}$ and $\boldsymbol{z_d}$ can independently reconstruct images, and that their combination ($\boldsymbol{x_u} + \alpha \boldsymbol{x_d}$) closely resembles the input image. The reconstructed image from $\boldsymbol{z_u}$ effectively reproduces the cortical folds of the brain, suggesting that $\boldsymbol{z_u}$ captures structural information that is domain-invariant. In contrast, the image reconstructed from $\boldsymbol{z_d}$ lacks detailed structural features and appears as a white haze, indicating that $\boldsymbol{z_d}$ encodes information unrelated to anatomical structures—presumably domain-specific characteristics. Although the image reconstructed from $\boldsymbol{z_u} + \boldsymbol{z_d}$ appears slightly blurred compared to the input, it still preserves the overall brain structure and cortical folds. These results suggest that combining domain-invariant and domain-specific components enables faithful reconstruction of brain images while enhancing the interpretability of domain-related information.

\subsection{Quantitative Results}
Table~\ref{tab:evaluation} summarizes the reconstruction error, disease classification performance, and domain classification performance for each model. Similar to previous studies, Noise and ComBat are methods that directly modify the extracted feature representations; therefore, evaluation on reconstructed images was not conducted for these models.

Based on RMSE and SSIM scores, the reconstruction capability of the proposed PL-SE-ADA model was slightly inferior to that of the Baseline and SE-ADA models. This can be attributed to PL-SE-ADA's training strategy, in which domain-specific information—assumed to be unrelated to brain structure—is added to the domain-invariant component in the image space. As a result, the proportion of non-structural domain information in the reconstructed brain image increases, likely leading to a decrease in scores that reflect the preservation of anatomical structures. 

Meanwhile, the macro F1 scores for disease classification indicate that the low-dimensional representations obtained by PL-SE-ADA retain more disease-relevant information than those from SE-ADA. This improvement is likely due to PL-SE-ADA's reconstruction strategy: whereas SE-ADA combines $\boldsymbol{z_u}$ and $\boldsymbol{z_d}$ in the latent space, PL-SE-ADA combines their corresponding image-space reconstructions $\boldsymbol{x_u}$ and $\boldsymbol{x_d}$, and computes reconstruction error with respect to the original input image. This training encourages $\boldsymbol{x_u}$ to align more closely with the final reconstructed image, thereby preserving structural features relevant to disease. These results suggest that PL-SE-ADA preserves brain structure more effectively in the image space, enhancing the representation of disease-specific features compared to SE-ADA.

\begin{figure*}[t]
  \centering
  \includegraphics[width=0.6\textwidth]{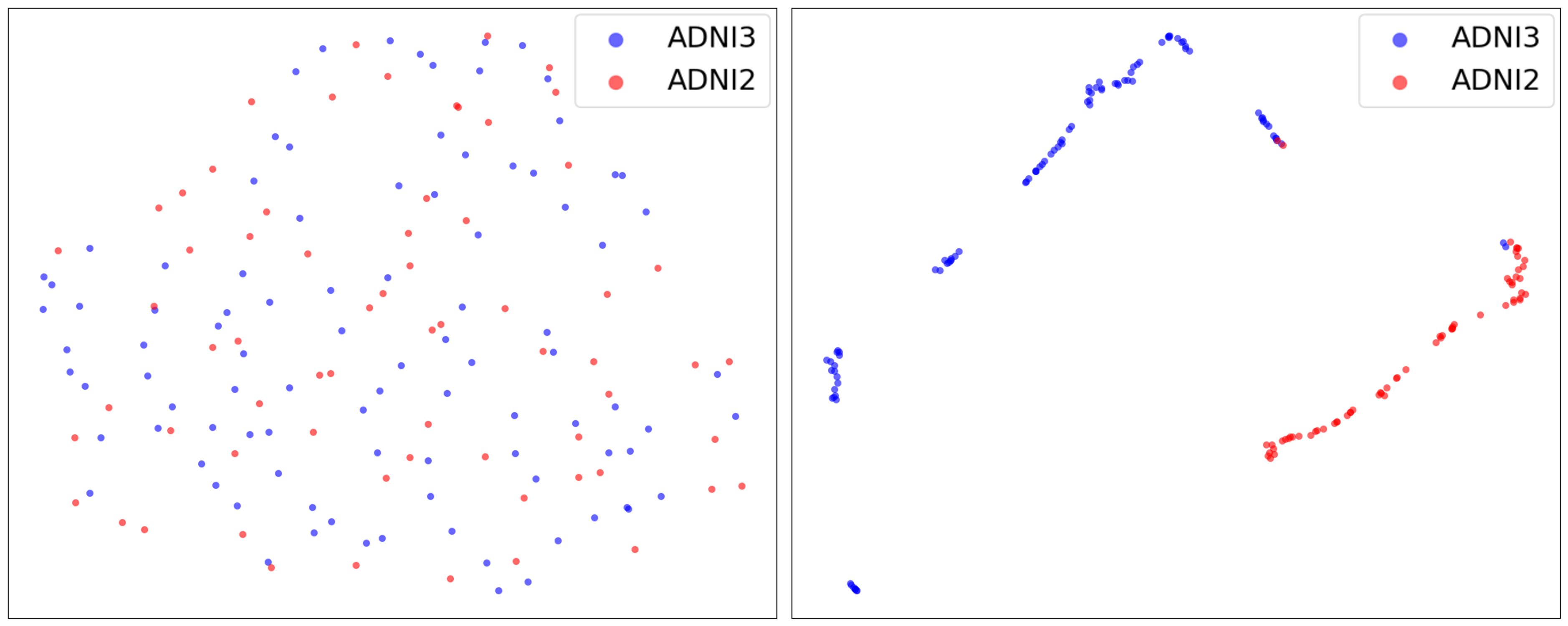}
  \caption{Visualization of domain-invariant biological representation ($\boldsymbol{z_u}$: left) and domain-dependent representation ($\boldsymbol{z_d}$: right) extracted by PL-SE-ADA and projected using UMAP.}
  \label{fig:UMAP}
\end{figure*}

The macro F1 score for domain classification was comparable between PL-SE-ADA and SE-ADA. Fig.~\ref{fig:UMAP} illustrates the two-dimensional distributions of $\boldsymbol{z_u}$ and $\boldsymbol{z_d}$, showing how these representations are separated across domains. As expected, $\boldsymbol{z_u}$, designed to be domain-invariant, exhibited no clear separation by domain, while $\boldsymbol{z_d}$, which encodes domain-specific information, showed distinct domain-wise clustering. These results suggest that the introduction of PL-SE-ADA does not compromise the original goal of achieving domain harmonization.

\begin{table*}[t]
\centering
\caption{Evaluation results for each model \\(reconstruction error, disease classification performance, domain classification performance).}
\label{tab:evaluation}
\resizebox{\textwidth}{!}{
\begin{tabular}{lccccccc}
\hline
\makecell{Model} & 
\makecell{RMSE ($\downarrow$)} & 
\makecell{SSIM ($\uparrow$)} & 
\makecell{Disease Classification\\F1 ($\uparrow$)} & 
\makecell{Domain Classification\\F1 ($\rightarrow 0.5$)} & 
\makecell{Availability of latent \\(z)}&
Availability of $\boldsymbol{z_d}$&
Interpretability \\
\hline
Baseline(CAE) & 0.0859 & 0.771 & 0.805 & 0.841&\cmark & &\\
+Noise & n/a & n/a & 0.489 & 0.532& \cmark&&\\
+Combat~\cite{johnson2007adjusting} & n/a & n/a & 0.825 & 0.572& &&\\
+ADA~\cite{dinsdale2021deep} & 0.0866 & 0.771 & 0.823 & 0.683&\cmark &&\\
+SE-ADA~\cite{10.1117/12.3046652} & \textbf{0.0858} & \textbf{0.774} & 0.828 & 0.522& \cmark&\cmark&\\
\textbf{+PL-SE-ADA} & 0.0991 & 0.729 & \textbf{0.875} & \textbf{0.512} & \cmark &\cmark &\cmark\\
\hline
\end{tabular}}
\end{table*}

\begin{table}[t]
\centering
\caption{Comparison of performance of PL-SE-ADA under varying domain ratios $\alpha$.}
\label{tab:evaluation_style}
\scalebox{0.72}{
\begin{tabular}{lcccc}
\hline
$\alpha$ & 
RMSE ($\downarrow$) &
SSIM ($\uparrow$) &
\makecell{Disease Classification\\F1 ($\uparrow$)} &
\makecell{Domain Classification\\F1 ($\rightarrow 0.5$)}\\
\hline
0.05 & 0.0983 & 0.731 & 0.859 & 0.602\\
0.1 & \textbf{0.0912} & \textbf{0.743} & 0.835 & 0.568\\
0.2 & 0.0991 & 0.729 & \textbf{0.875} & \textbf{0.512}\\
0.5 & 0.1082 & 0.721 & 0.838 & 0.621\\
1.0 & 0.1453 & 0.653 & 0.831 & 0.631 \\
1.5 & 0.1498 & 0.624 & 0.814 & 0.576 \\
\hline
\end{tabular}}
\end{table}

\begin{figure}[t]
  \centering
  \includegraphics[width=0.35\textwidth]{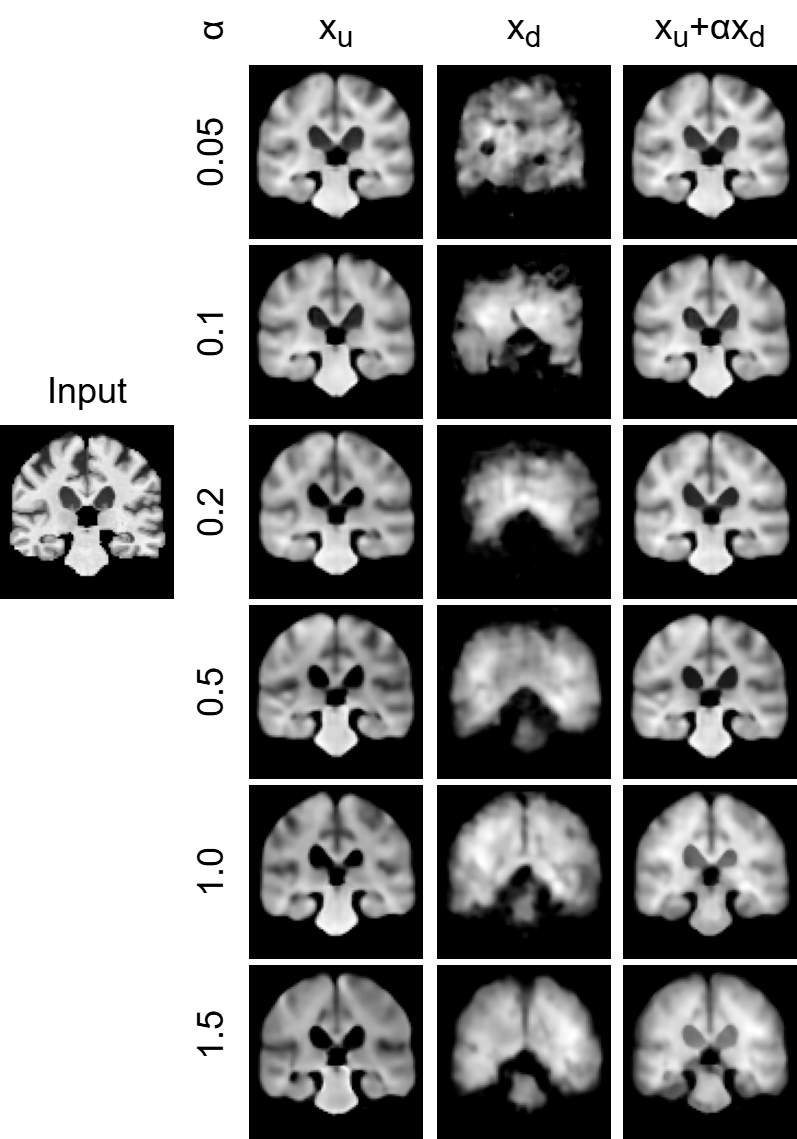}
  \caption{Comparison of image reconstruction under varying domain ratios $\alpha$.}
  \label{fig:stylehiritu}
\end{figure}

\subsection{Impact of hyperparameter}
Fig.~\ref{fig:stylehiritu} visualizes the reconstructed images obtained by varying the weighting parameter $\alpha$, while Table~\ref{tab:evaluation_style} summarizes the reconstruction error, disease classification accuracy, and domain classification accuracy for each value of $\alpha$. As $\alpha$ increases, the reconstructed images become progressively more whitish, with the regions derived from $\boldsymbol{z_d}$ expanding. This observation suggests that increasing the contribution of domain-specific information during reconstruction results in visual content that becomes more localized to domain-related variations. By comparing reconstruction accuracy, disease classification accuracy, and domain classification accuracy, it is considered that $\alpha = 0.2$ provides the best balance between overall performance and interpretability.

\section{Further applications and limitations}
This study demonstrated the strengths of PL-SE-ADA in domain alignment and visual interpretability. Its ability to handle both biological and domain-specific features makes it promising for applications like style transfer and domain adaptation. 

However, the current evaluation used a limited dataset with few domains. Future work will focus on improving image reconstruction quality, hyperparameter tuning, and broader experiments.
\section{Conclusion}
In this paper, we propose PL-SE-ADA, an extension of SE-ADA, that disentangles domain-specific and biological signals with visual interpretability. Evaluated on the ADNI2/3 dataset, it achieves competitive or better performance in data preservation and domain alignment, while enhancing interpretability-crucial in medical applications. By treating both domain and disease-specific variation as style information, our method also supports potential applications in style conversion and manipulation.
{
\section*{Acknowledgments}
\small
\begin{spacing}{0.9}
This research was supported in part by the Ministry of Education, Science, Sports and Culture of Japan (JSPS
KAKENHI), Grant-in-Aid for Scientific Research (C), 21K12656 (2021-2023) and 24K15706 (2024-2026).
Data collection and sharing for this project was funded by the Alzheimer’s Disease Neuroimaging Initiative
(ADNI) (National Institutes of Health Grant U01 AG024904) and DOD ADNI (Department of Defense award
number W81XWH-12-2-0012). ADNI is funded by the National Institute on Aging, the National Institute
of Biomedical Imaging and Bioengineering.
\end{spacing}
}

\bibliographystyle{IEEEtran}
\bibliography{references}

\end{document}